\documentclass[wcp]{jmlr}

\usepackage{longtable}
\usepackage{booktabs}
\usepackage{amsfonts}
\usepackage{algorithm}
\usepackage{algorithmic}
\usepackage{arydshln}
\usepackage{listings}
\usepackage{xcolor}

\lstdefinestyle{mystyle}{
    backgroundcolor=\color{white},   
    commentstyle=\color{green},
    keywordstyle=\color{magenta},
    numberstyle=\tiny\color{gray},
    stringstyle=\color{blue},
    basicstyle=\ttfamily\footnotesize,
    breakatwhitespace=false,         
    breaklines=true,                 
    captionpos=b,                    
    keepspaces=true,                 
    numbers=left,                    
    numbersep=5pt,                  
    showspaces=false,                
    showstringspaces=false,
    showtabs=false,                  
    tabsize=2,
}

\makeatletter
\let\Ginclude@graphics\@org@Ginclude@graphics 
\makeatother

\title[Privileged Zero-Shot AutoML]{Privileged Zero-Shot AutoML}

  \author{\Name{Nikhil Singh} \Email{nsingh1@mit.edu}
   \addr MIT \AND
  \Name{Brandon Kates} \Email{bjk224@cornell.edu}
   \addr Cornell University \AND
  \Name{Jeff Mentch} \Email{jsmentch@mit.edu}
   \addr Harvard University \AND
  \Name{Anant Kharkar} \Email{agk2151@columbia.edu}
   \addr Columbia University \AND
   \Name{Madeleine Udell} \Email{udell@cornell.edu}
   \addr Cornell University \AND
   \Name{Iddo Drori} \Email{idrori@mit.edu}
   \addr MIT}

\begin{document}

\maketitle

\begin{abstract}
This work improves the quality of automated machine learning (AutoML) systems by using dataset and function descriptions while significantly decreasing computation time from minutes to milliseconds by using a zero-shot approach. Given a new dataset and a well-defined machine learning task, humans begin by reading a description of the dataset and documentation for the algorithms to be used. This work is the first to use these textual descriptions, which we call privileged information, for AutoML. We use a pre-trained Transformer model to process the privileged text and demonstrate that using this information improves AutoML performance. Thus, our approach leverages the progress of unsupervised representation learning in natural language processing to provide a significant boost to AutoML. We demonstrate that using only textual descriptions of the data and functions achieves reasonable classification performance, and adding textual descriptions to data meta-features improves classification across tabular datasets. To achieve zero-shot AutoML we train a graph neural network with these description embeddings and the data meta-features. Each node represents a training dataset, which we use to predict the best machine learning pipeline for a new test dataset in a zero-shot fashion. Our zero-shot approach rapidly predicts a high-quality pipeline for a supervised learning task and dataset. In contrast, most AutoML systems require tens or hundreds of pipeline evaluations. We show that zero-shot AutoML reduces running and prediction times from minutes to milliseconds, consistently across datasets. By speeding up AutoML by orders of magnitude this work demonstrates real-time AutoML.
\end{abstract}


\section{Introduction}
\label{sec:introduction}

\begin{figure}
    \centering
    \includegraphics[width=\textwidth]{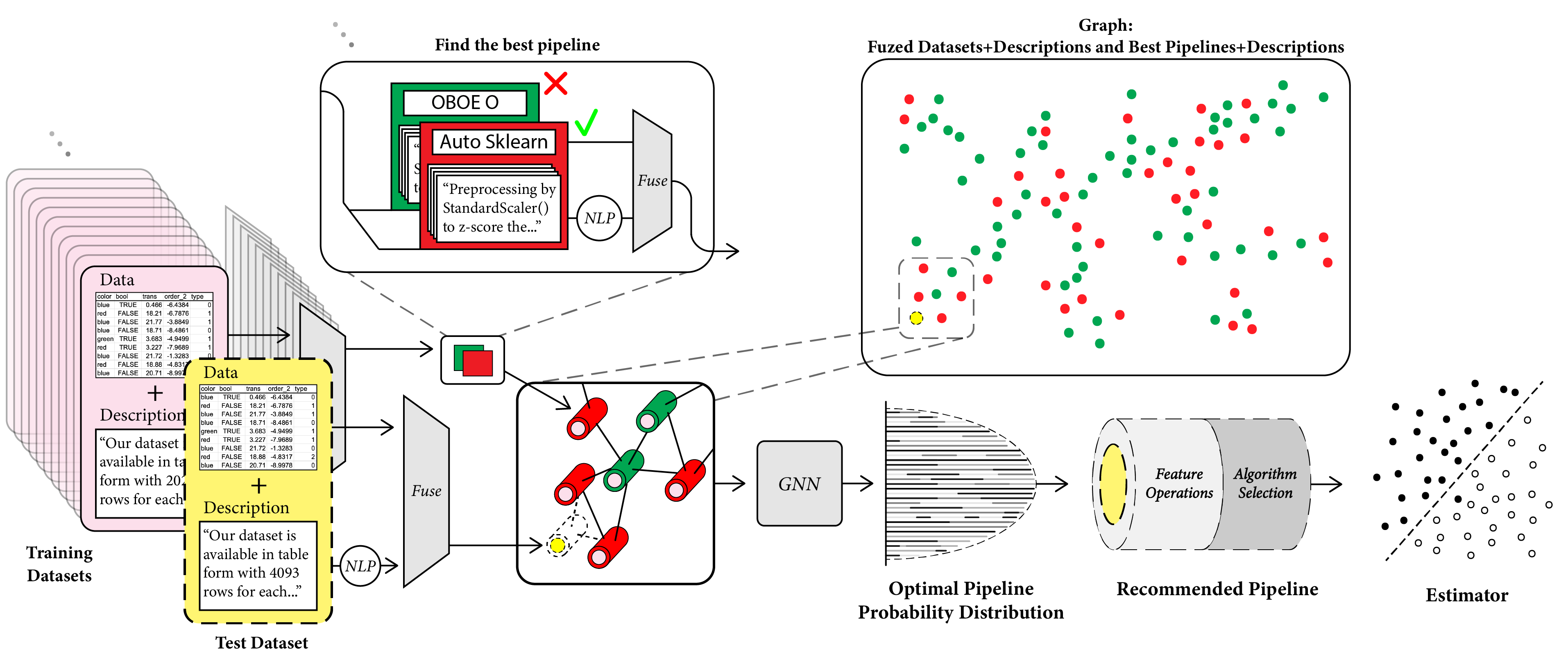}
    \caption{Overview of our method. We leverage dataset descriptions and other AutoML methods to provide zero-shot ML pipeline selection.}
    \label{fig:architecture}
\end{figure}

A data scientist facing a challenging new supervised learning task does not generally invent a new algorithm. Instead, they consider what they know about the dataset and which algorithms have worked well for similar datasets in the past. Automated machine learning (AutoML) seeks to automate such tasks, enabling the widespread and accessible use of machine learning by non-experts. A major challenge in the field is to develop fast, efficient algorithms to accelerate machine learning applications \citep{kokiopoulou2019fast}.

This work develops automated solutions that exploit human expertise to learn which datasets are similar and which algorithms perform best. We use a transformer-based language model \citep{vaswani2017attention} to process text descriptions of datasets and algorithms, and a feature extractor \citep{byu2018} to represent the data itself. Our approach fuses each of these representations, representing each dataset as a node in a graph of datasets. We train our model on other existing AutoML system solutions, specifically: AutoSklearn \citep{feurer2015efficient} and OBOE \citep{yang2019oboe}. By leveraging these existing systems and openly accessible datasets, we achieve state-of-the-art results using multiple approaches across diverse classification problems.

To predict a machine learning pipeline, a simple idea is to use a pipeline that performed well on the same task and on similar datasets; however, what constitutes a similar dataset? The success of an AutoML system often hinges on this question, and different frameworks have different answers: for example, AutoSklearn \citep{feurer2015efficient} computes a set of meta-features, features describing the data features, for each dataset, while OBOE \citep{yang2019oboe} uses the performance of a few fast, informative models to compute latent features. More generally, for any supervised learning task, one can view the list of recommended algorithms generated by any AutoML system as a vector describing that task. This work is the first to use the information that a human would check first: a summary description of the dataset and algorithms, written in free text. These dataset features induce a metric structure on the space of datasets. Under an ideal metric, a model that performs well on one dataset would also perform well on nearby datasets. The methods we develop in this work show how to learn such a metric using the recommendations of an AutoML framework together with the dataset description. We provide a new zero-shot AutoML method that predicts accurate machine learning pipelines for an unseen dataset and classification task in real-time.

\paragraph{Leveraging advances in NLP.} Bringing techniques from NLP to AutoML, we specifically use a large-scale Transformer model to extract information from the description of both the datasets and algorithms. This allows us to access large amounts of relevant information that existing AutoML systems are typically not privy to. These embeddings of dataset and pipeline descriptions are fused with data meta-features to build a graph where each dataset is a single node. This graph is then the input to a graph neural network (GNN).

\paragraph{Real-time.} Given a new dataset, our real-time AutoML method predicts a pipeline with good performance within milliseconds. The running time of this predicted pipeline is typically up to a second, mainly for hyperparameter tuning. The accuracy of our method is competitive with state-of-the-art AutoML methods that are given minutes, thus, reducing computation time by orders of magnitude while improving performance.

\paragraph{GNN architecture.} Generally, graph neural networks are used for three main tasks: (i) node prediction, (ii) link prediction, and (iii) sub-graph or entire graph property prediction. In this work we use a GNN for node prediction, predicting the best machine learning pipeline for an unseen dataset. Specifically, we use a graph attention network (GAT) \citep{velivckovic2018gat} with neighborhood aggregation, in which an attention function adaptively controls the contribution of neighbors. An advantage of using a GNN in our use case is that data, metadata, and algorithm information is shared between datasets (graph nodes) by messages passed between the nodes of the graph. In addition, GNNs generalize well to new unknown datasets using their aggregated weights learned during training which are shared with the test dataset during testing. Beyond just a single new dataset, GNNs can generalize further to an entire new set of datasets.

\paragraph{Leveraging existing AutoML systems.}
This work demonstrates how solutions from existing AutoML systems are used to train a new AutoML model. Our flexible architecture can be extended to use pipeline recommendations from any number of other AutoML systems to further improve performance.

\section{Related Work}\label{sec:work}

AutoML is an emerging field of machine learning with the potential to transform the practice of data science by automatically choosing a model to best fit the data. Several comprehensive surveys of the field are available \citep{he2019automl,zoller2019survey}. The most straightforward approach to AutoML considers each dataset in isolation and asks how to choose the best hyperparameter settings for a given algorithm. While the most popular method is still grid search, other more efficient approaches include Bayesian optimization \citep{snoek2012} and random search \citep{solis1981minimization}. Recommender systems learn, often exhaustively, which algorithms and hyperparameter settings perform best for a training set-of-datasets and use this information to select better algorithms on a test set without exhaustive search. This approach reduces the time required to find a good model. An example is OBOE \citep{yang2019oboe} and TensorOBOE \citep{yang2020tensor}, which fit a low rank model to learn the low-dimensional representations for the models or pipelines and datasets that best predict the cross-validated errors, among all bi-linear models. To find promising models for a new dataset, OBOE runs a set of fast but informative algorithms on the new dataset and uses their cross-validated errors to infer the feature vector for the new dataset. A related approach \citep{fusi2018probabilistic} using probabilistic matrix factorization powers Microsoft Azure's AutoML service \citep{azure}.

Auto-Tuned Models \citep{atm2017} represent the search space as a tree with nodes being algorithms or hyperparameters and searches for the best branch using a multi-armed bandit. AlphaD3M \citep{drori2018alphad3m,drori2019alphad3m} formulates AutoML as a single player game. The system uses reinforcement learning with self-play and a pre-trained model which generalizes from many different datasets and similar tasks.

TPOT \citep{olson2019tpot} and Autostacker \citep{chen2018autostacker} use genetic programming to choose both hyperparameter settings and a topology of a machine learning pipeline. TPOT represents pipelines as trees, whereas Autostacker represents them as layers.

AutoSklearn \citep{feurer2015efficient} chooses a model for a new dataset by first computing data meta-features to find nearest-neighbor datasets. The best-performing methods on the neighbors are refined by Bayesian optimization and used to form an ensemble. End-to-end learning of machine learning pipelines can be performed using differentiable primitives \citep{mitar2017} forming a directed acyclic graph. One major factor in the performance of an AutoML system is the base set of algorithms it can use to compose more complex pipelines. For a fair comparison, in our numerical experiments we compare our proposed methods only to other AutoML systems that use Scikit-learn \citep{scikit-learn} primitives.

\paragraph{Embeddings.}
Language has a common unstructured representation as a sequence of words, sentences, or paragraphs. The most significant recent progress in NLP is large-scale Transformers \citep{devlin2018bert,shoeybi2019megatron,raffel2019exploring} based on attention mechanisms \citep{vaswani2017attention}. An unsupervised corpus of text is transformed into a supervised dataset by defining content-target pairs along the entire text: for example, target words that appear in each sentence, or target sentences which appear in each paragraph. A language model is first trained to learn a low dimensional embedding of words or sentences followed by a map from low dimensional content to target \citep{mikolov2013efficient}. This embedding is then used on a new, unseen and small dataset in the same low-dimensional space. Our work applies such embeddings to AutoML. In a similar fashion to recent work \citep{drori2019embedding} we use an embedding for the dataset and algorithm descriptions, however, in this work we model the non-linear interactions between these embedding using a neural network.

\section{Methods}\label{sec:section1}

\begin{table*}
\small
\centering
\scriptsize
\begin{tabular}{ll}
 \textbf{Notation} & \textbf{Description} \\
 \hline
 $\mathcal{D}$ & Dataset\\
 $\mathcal{M(D)}$ & Dataset description\\
 \hline
 $\mathcal{P}$ & Machine learning pipeline \\
 $\mathcal{M(P)}$ & Machine learning pipeline description \\
 $\textrm{C} \in \textrm{O, S}$ & TensorOboe, AutoSklearn \\
 $\mathcal{P_\textrm{C}(D)}$ & Pipeline recommended by $\textrm{C}$ on dataset $\mathcal{D}$ \\
 $\mathcal{P}_{\star}(\mathcal{D})$ & Best pipeline on dataset $\mathcal{D}$\\
 $\hat{\mathcal{P}}(\mathcal{D})$ & Predicted pipeline on dataset $\mathcal{D}$\\
 $\mathcal{R(P, D)}$ & Performance of running pipeline $\mathcal{P}$ on dataset $\mathcal{D}$\\
 \hline
 $\mathcal{F_{D}}$ & Data meta-features\\
 $\mathcal{F_{M}} = E(\mathcal{M(D)})$ & Embedding of dataset description\\
 $\mathcal{F_{D,M}} = [\mathcal{F_{D}},\mathcal{F_{M}}]$ & Concatenation\\
 $\mathcal{F_{P}} = E(\mathcal{M(P)})$ & Embedding of pipeline description\\
 \hline
 $\mathcal{G}$ & Datasets graph\\
 $i \in V$ & Node in $\mathcal{G}$\\
 $j \in \mathcal{N}(i)$ & Neighbors $j$ of node $i$\\
 $\mathcal{F}_{i} = f_{\phi}(\mathcal{F}_{\mathcal{D}_{i},\mathcal{M}_{i}})$ & Fusion network output on graph node\\
 $\mathbf{u}_{i} = g_{\theta}(\mathbf{v}_{i})$ & Fusion network, features of node in GNN\\
 $\{\mathbf{u}_{j}\}_{j \in \mathcal{N}(i)}$ & Features of node neighbors in GNN\\
 $h_{W,z}(\mathbf{u}_{i},\{\mathbf{u}_{j}\}_{j \in \mathcal{N}(i)})$ & GNN with parameters $W,z$\\
 \hline
\end{tabular}
    \caption{AutoML notation and descriptions.}
    \label{tab:notation}
\end{table*}

Our zero-shot AutoML predicts a machine learning pipeline for a classification task on a dataset based on the dataset description and data, and based on other datasets, their relationships, and their recommended pipelines by AutoML systems. We embed the dataset description and extract data meta-features to construct a graph of datasets where each node represents a dataset. The graph is processed using a graph neural network (GNN). Each node of the graph contains a feature vector which is a learned representation of the data, its description, and the description of its best predicted pipeline. This is obtained through two successive fusion networks: the fusion of the description embedding and data meta-features, which is in turn fused with the embedding of the best AutoML solution descriptions (with feature processing and estimation component embeddings concatenated). The machine learning pipeline for a new dataset is predicted by the GNN. A detailed architecture is illustrated in Fig. \ref{fig:architecture} and described by Algorithms \ref{alg:zero-shot-automl-training} and \ref{alg:zero-shot-automl-testing}. The notation used in this work is detailed in Table \ref{tab:notation}.

\subsection{Pre-processing}
Our pre-processing consists of (i) dataset description embedding; (ii) dataset meta-feature extraction; and (iii) pipeline computation and description embeddings, as described next and summarized in Algorithm \ref{alg:zero-shot-automl-preprocessing}.

\vspace{-4pt}

\paragraph{Dataset Description Embedding.}
We create a feature vector by embedding the description $\mathcal{M(D)}$ of each dataset as a $1024$-dimensional vector $\mathcal{F}_{\mathcal{M}}=E(\mathcal{M(D)})\in\mathbb{R}^{1024}$ using T5 \citep{raffel2019exploring}. Example dataset descriptions include the following (abridged for space reasons):

\begin{quote}
    "…The dataset consist of several assignments. Each assignment consists of a question followed by ten sentences (titles of news articles). One of the sentences is the correct answer to the question (C) and five of the sentences are irrelevant to the question (I). Four of the sentences are relevant to the question (R), but they do not answer it…The first column is the line number, second the assignment number and the next 22 columns (3 to 24) are the different features. Columns 25 to 27 contain extra information about the example…The 22 features provided are commonly used in psychological studies on eye movement. All of them are not necessarily relevant in this context…"
\end{quote}

We observe that this description contains rich contextual information that a data scientist designing a pipeline is likely to find useful. To survey one other example:

\begin{quote}
    "Predict a biological response of molecules from their chemical properties. Each row in this data set represents a molecule. The first column contains experimental data describing an actual biological response; the molecule was seen to elicit this response (1), or not (0). The remaining columns represent molecular descriptors (d1 through d1776), these are calculated properties that can capture some of the characteristics of the molecule  for example size, shape, or elemental constitution. The descriptor matrix has been normalized. The original training and test set were merged."
\end{quote}

Again, this description encodes and makes available helpful domain knowledge. While descriptions in our set-of-datasets are not uniformly detailed or context-rich, they introduce additional knowledge into the process that we can benefit from if we adequately represent and incorporate it.

\vspace{-4pt}

\paragraph{Data meta-features.}
We compute meta-features $\mathcal{F_{D}}\in\mathbb{R}^{149}$ for the dataset $\mathcal D$ using a feature extractor \citep{byu2018}, restricting to meta-features that are computed in less than a second on datasets used in our experiments. Meta-features include statistics of the datasets and results of simple algorithms.

\vspace{-4pt}

\paragraph{Pipelines and pipeline embedding.}
For each dataset, we compute the recommended pipeline returned by 
AutoML systems TensorOBOE (O) and AutoSklearn (S). We create feature vectors for recommended pipelines by embedding the Scikit-learn documentations for pre-processor or feature selector and estimator (which is unique within each pipeline). Again, we use the T5 model to form a $2048$-dimensional embedding:

\begin{equation}
E(\mathcal{M(P_{C}(D))})\in\mathbb{R}^{2048} = E(\mathcal{M(P^\mathrm{feat}_{C}(D))})\in\mathbb{R}^{1024} \oplus E(\mathcal{M(P^\mathrm{est}_{C}(D))})\in\mathbb{R}^{1024}
\end{equation}

for each pipeline, where $C$ ranges over the AutoML methods O and S, $\mathcal P^\mathrm{feat}$ is the feature processor description, $\mathcal P^\mathrm{est}$ is the estimator description, and $\oplus$ represents concatenation. The best-performing pipeline $\mathcal P^\star$ returned by any AutoML system serves as our training label: we train our system to recommend this pipeline. Available options for feature processing and estimation are enumerated in Table \ref{tab:algorithms}.

\begin{table}[!h]
\small
\centering
\begin{tabular}{l|l}
\textbf{Estimation} &                        \textbf{Feature Processing} \\
\hline
random\_forest&                       \textbf{Scaling or Normalization}\\
bagging &                              standardscaler \\
decision\_tree &                    robustscaler \\
liblinear\_svc &                    minmaxscaler \\  
gradient\_boosting &  normalizer \\                         
libsvm\_svc & maxabsscaler\\                              
extra\_trees & \textbf{Dimensionality Reduction}\\                               
bernoulli\_nb & pca\\              
adaboost & fastica\\                       
k\_nearest\_neighbors & \textbf{Feature Extraction}\\                      
multinomial\_nb & polynomial\\                      
passive\_aggressive & rbfsampler\\                               
gaussian\_nb & \textbf{Feature Selection}\\                        
logisticregression & selectfwe \\                      
sgd & variancethreshold \\                        
qda & selectfrommodel \\                
lda & select\_percentile\_classification \\                  
xgbclassifier & rfe\\
\hline
\end{tabular}
\caption{Scikit-learn machine learning primitives: estimators and feature processors available to our AutoML system.}
\label{tab:algorithms}
\end{table}

\vspace{-4pt}

\paragraph{Fused dataset representations.}
The combined representation of dataset $\mathcal{D}_{i}$ with description $\mathcal{M}(\mathcal{D}_{i})$ fuses together the dataset description embedding and data meta-features using a neural network:

\begin{equation}
\mathcal{F}_{i} = f_{\phi}([\mathcal{F}_{\mathcal{D}_{i}},\mathcal{F}_{\mathcal{M}_{i}}])\in\mathbb{R}^{512}.
\end{equation}

We also represent the dataset and its best pipeline by fusing this representation with the pipeline embedding using a second neural network:

\begin{equation}
\mathbf{u}_i=g_{\theta}([\mathcal{F}_{i}, \mathcal{F}_{\mathcal{P_{\star}}(\mathcal{D}_{i})}])\in\mathbb{R}^{512}
\end{equation}
 
\begin{figure*}
\centering
\includegraphics[width=0.9\textwidth]{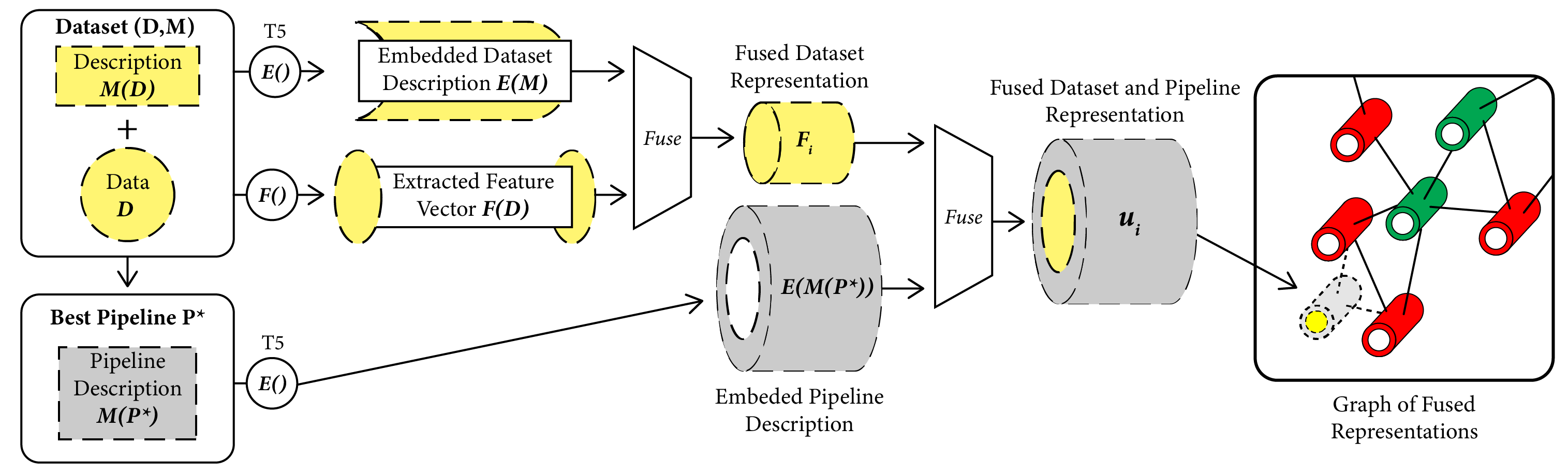}
\caption{AutoML fusion networks and dataset graph construction. Each dataset and corresponding description is paired with the best predicted pipeline (by either TensorObBOE or AutoSklearn). The dataset meta-features and description embedding are fused by a neural network, and then subsequently this representation is fused with the pipeline embedding through another neural network. The resulting representations then become nodes in a kNN graph.}
\label{fig:fusion}
\end{figure*}

These fused representations improve performance compared to concatenation. An illustration of the fusion networks is provided in Fig. \ref{fig:fusion}.

\begin{algorithm}
\small
   \caption{Zero-shot AutoML pre-processing}
   \label{alg:zero-shot-automl-preprocessing}
\begin{algorithmic}
   \STATE {\bfseries Input:} training datasets $\{(\mathcal{D}_i, \mathcal{M}_i)\}_{i \in V}$. 
   \STATE {\bfseries Output:} features $\{\mathcal{F_{M}}_i, \mathcal{F_{D}}_i, \mathcal{F}_{\mathcal{P}_{\star}(\mathcal{D}_i)}\}_{i \in V}$.
   
   \FOR{$i=1$ {\bfseries to} $n$}
     \STATE compute embedding of description $\mathcal{F_{M}}_i = E(\mathcal{M}_i)$
     \STATE compute data meta-features $\mathcal{F_{D}}_i$
     \FORALL {$\textrm{C} \in \textrm{O, S}$}
       \STATE compute recommended pipeline $\mathcal{P}_\textrm{C}(\mathcal{D}_i)$
       \STATE compute performance on dataset $\mathcal{R}(\mathcal{P}_\textrm{C}, \mathcal{D}_i)$
     \ENDFOR
     \STATE select best performing pipeline $\mathcal{P}_{\star}(\mathcal{D}_i)$
     \STATE embed pipeline $\mathcal{F}_{\mathcal{P}_{\star}(\mathcal{D}_i)} = E(\mathcal{M}(\mathcal{P}_{\star}(\mathcal{D}_i)))$
   \ENDFOR
\end{algorithmic}
\end{algorithm}

\subsection{Graph Representation}
\label{sec:graph-representation}
We build a graph $\mathcal{G}=(V,E)$ where each node $i \in V$ represents the dataset $\mathcal D_i$ and has feature vector $\mathbf{u}_i$. The graph is defined by nodes and edges:

\vspace{-4pt}

\paragraph{Nodes.} As noted, the feature vector $\mathbf{u}_i=g_{\theta}([\mathcal{F}_{i}, \mathcal{F}_{\mathcal{P_{\star}}(\mathcal{D}_{i})}])\in\mathbb{R}^{512}$ for node $i$ representing dataset $\mathcal{D}_{i}$ with description $\mathcal{M}(\mathcal{D}_{i})$ combines the fused dataset representation $\mathcal{F}_{i}\in\mathbb{R}^{512}$ and the pipeline embedding $\mathcal{F}_{\mathcal{P}}=E(\mathcal{P_{\star}})\in\mathbb{R}^{1024}$ for the pipeline $\mathcal{P}_{\star}$ that performed best on the dataset, through a second fusion network $g_{\theta}$. During training, we mask the pipeline embedding from the feature vector and learn to predict a node using the GNN.

\vspace{-4pt}

\paragraph{Edges.} To compute the edges of the graph $\mathcal{G}$, we compute the distance $d$ between each pair of datasets $i,j$ as $d = \|\mathcal{F}_{i} - \mathcal{F}_{j}\|_{2}$ where $\mathcal{F}_{i}$ and $\mathcal F_j$ are the fused dataset representations for the datasets. Two datasets are connected by an edge if dataset $j$ is one of the $k$ nearest neighbors of dataset $i$ or vice versa. In our experiments, we chose $k=20$ and found that our method is reasonably robust to the choice of $k$, and that a k-NN graph outperforms a threshold-based graph.

At training time, we build the graph on the training datasets. At test time, given a new test dataset, we dynamically connect the new node to the graph using its fused feature representation $f_{\phi}([\mathcal{F}_{\mathcal{M}_\textrm{test}}, \mathcal{F}_{\mathcal{D}_\textrm{test}}])$ to choose edges. The edges for the new dataset are rapidly chosen without fitting a single machine learning model.

\subsection{Neural Network Architecture}
The neural networks we train for zero-shot AutoML consist of two fusion networks and a graph attention network which is a type of GNN. The fusion networks are used to capture the non-linear interactions between the features corresponding to the descriptions, the data meta-features, and the pipeline embedding. The GNN predicts the best pipeline for a new dataset based on the weights optimized during training as described next.

\paragraph{Graph Attention Network.}
A graph attention network (GAT) \citep{velivckovic2018gat} with two parallel fully-connected output layers is used to predict the best pipeline for a new dataset. Each layer $l=1,...,L$ of the GNN updates the feature vector at the $i$-th node as:

\begin{equation}
\mathbf{u}_i^{l} =\alpha_{ii}W\mathbf{u}_i^{l-1} +\sum_{j\in\mathcal{N}(i)}\alpha_{ij}W\mathbf{u}_j^{l-1},
\end{equation}

where $W$ is a learnable weight matrix, $\mathcal{N}(i)$ are the neighbors of the $i$-th node, and $\alpha_{ij}$ are the attention coefficients, defined as:

\begin{equation}
\alpha_{ij} = \frac{\exp(\sigma(z^{\top}\left[W\mathbf{u}_i,W\mathbf{u}_j\right]))}{\sum_{k\in\mathcal{N}(i)}\exp(\sigma(z^{\top}\left[W\mathbf{u}_i,W\mathbf{u}_k\right]))},
\end{equation}

where $z$ is a learnable vector, and $\sigma(\cdot)$ is the leaky ReLU activation function. 

Our GNN consists of 3 GAT layers followed by two parallel fully-connected layers. Applying a softmax to each of these computes a vector of probabilities over estimators and feature-processors respectively. Hence the output of the GAT is a probability distribution over pipelines for each node. The network recommends the pipeline that maximizes this probability. Alternatively, it is possible to sample from this probability distribution to obtain several pipelines that are combined into an ensemble.

\subsection{Training and Testing}
\paragraph{Training.}
We train on a set of 165 datasets, largely taken from OpenML \footnote{\href{https://openml.org/}{https://openml.org/}}. Our training process is described in Algorithm \ref{alg:zero-shot-automl-training}. At each training iteration, we randomly select a node $i$. We mask the pipeline embedding of the $i$-th node as $\mathbf{u}_i=g_{\theta}([\mathcal{F}_i,\mathbf{0}])$. The true label is defined as the pipeline with best performance among the two AutoML systems $\mathcal{P}_{\star}(\mathcal{D}_i)$ on the $i$-th dataset. The resulting problem consists of two multi-class classification problems with as many classes as there are distinct feature processors and estimators respectively.

The loss function is defined by the sum of two cross-entropies, one for the feature processor and one for the estimator, between the probability $\hat{\mathbf{p}}$ of predicted pipeline $\hat{\mathcal{P}}(\mathcal{D}_{i})$ and one-hot encoding $\mathbf{y}$ of the best pipeline $\mathcal{P}{\star}(\mathcal{D}_i)$:

\begin{equation}
\mathcal{L}(\hat{\mathcal{P}}(\mathcal{D}_i),\mathcal{P}_{\star}(\mathcal{D}_i)) = \sum_{c \in \{\mathrm{feat}, \mathrm{est}\}} \mathcal{L}(\hat{\mathcal{P}^c}(\mathcal{D}_i),\mathcal{P}^c_{\star}(\mathcal{D}_i)) = -\sum_{l=1}^{m}\mathbf{y}_l\log(\hat{\mathbf{p}}_l)
\end{equation}

\begin{algorithm}
\small
   \caption{Privileged zero-shot AutoML training}
   \label{alg:zero-shot-automl-training}
\begin{algorithmic}
   \STATE {\bfseries Input:} training datasets, descriptions $\{\mathcal{D}_i, \mathcal{M}(\mathcal{D}_i)\}_{i \in V}$. 
   \STATE {\bfseries Output:} datasets graph $\mathcal{G}$, GNN $h_{W,z}$, fusion networks $f_{\phi}$ and $g_{\theta}$.
   
   \STATE pre-process: compute $\{\mathcal{F_{M}}_i, \mathcal{F_{D}}_i, \mathcal{F}_{\mathcal{P}_{\star}(\mathcal{D}_i)}\}_{i \in V}$
   
   \STATE initialize fusion networks weights $\phi, \theta$.
   \STATE initialize GNN weights $W, z$.
   \FOR {each backprop iteration}
     \STATE generate updated datasets graph $\mathcal{G}$:
     \FOR{$i=1$ {\bfseries to} $n$}
       \STATE compute fused representation $\mathcal{F}_i = f_{\phi}(\mathcal F_{\mathcal{D}_i, \mathcal{M}_i})$
     \ENDFOR
     \STATE compute pairwise distances $d(\mathcal{F}_i,\mathcal{F}_j)_{i,j \in V}$
     \FOR{$i=1$ {\bfseries to} $n$}
       \STATE connect node $i$ to k-NN nodes $\mathcal{N}(i)$
     \ENDFOR
     
     \STATE select random node $i$ in $\mathcal{G}$
     \STATE compute $\mathbf{u}_{i} = g_{\theta}(\mathcal{F}_i,\mathbf{0})$
     \FORALL {$j \neq i$}
       \STATE compute $\mathbf{u}_{j} = g_{\theta}(\mathcal{F}_j,\mathcal{F}_{\mathcal{P}_{*}(\mathcal{D}_j)})$
     \ENDFOR
     \STATE predict best pipeline components $\hat{\mathcal{P}}(\mathcal{D}_i) = h_{W,z}(\mathbf{u}_i,\{\mathbf{u}_j\}_{j \in \mathcal{N}(i)})$
     \STATE compute loss $\mathcal{L}(\hat{\mathcal{P}}(\mathcal{D}_i),\mathcal{P}_{\star}(\mathcal{D}_i))$
     \STATE update weights
   \ENDFOR
\end{algorithmic}
\end{algorithm}

\paragraph{Testing.}
\begin{algorithm}[!h]
\small
   \caption{Privileged zero-shot AutoML testing}
   \label{alg:zero-shot-automl-testing}
\begin{algorithmic}
   \STATE {\bfseries Input:} dataset $\mathcal{D}_{i}$, description $\mathcal{M}(\mathcal{D}_{i})$, datasets graph $\mathcal{G}$, GNN, s.t. $i \not\in V$ (disjoint train and test).
   \STATE {\bfseries Output:} predict best pipeline components $\hat{\mathcal{P}}(\mathcal{D}_{i})$ for task on dataset.
   \STATE generate new node $i$ in $\mathcal{G}$:
   \STATE compute embedding of description $\mathcal{F_{M}} = E(\mathcal{M}(\mathcal{D}_{i}))$
   \STATE compute data meta-features $\mathcal{F_{D}}$
   \STATE compute fused representation $\mathcal F = f_{\phi}(\mathcal F_\mathcal{D}, \mathcal F_{\mathcal{M}})$
   \STATE connect node $i$ to k-NN nodes $j \in \mathcal{N}(i)$, $V = V \cup \{i\}$.
   \STATE compute $\mathbf{u}_{i} = g_{\theta}(\mathcal{F},\mathbf{0})$
   \STATE predict best pipeline components $\hat{\mathcal{P}}(\mathcal{D}_{i}) = h_{W,z}(\mathbf{u}_i,\{\mathbf{u}_j\}_{j \in \mathcal{N}(i)})$
\end{algorithmic}
\end{algorithm}

Our testing process is described in Algorithm \ref{alg:zero-shot-automl-testing}. 
Given a new dataset $\mathcal{D}$ and description $\mathcal{M}$, 
we compute the description embedding $\mathcal{F}_{\mathcal{M}}$ and data meta-features $\mathcal{F}_{\mathcal{D}}$ and the fused dataset representation $\mathcal F$. We use this representation to compute the edges of this new node in the graph of all datasets.
Next, we add the new node, with features $\mathbf{u} = g_\theta([\mathcal F, \mathbf{0}])$, to the current graph, replacing the embedding of the pipeline with the zero vector. Finally, we use the graph neural network to recommend a pipeline for the test dataset.

\begin{figure}
    \centering
    \includegraphics[width=\textwidth]{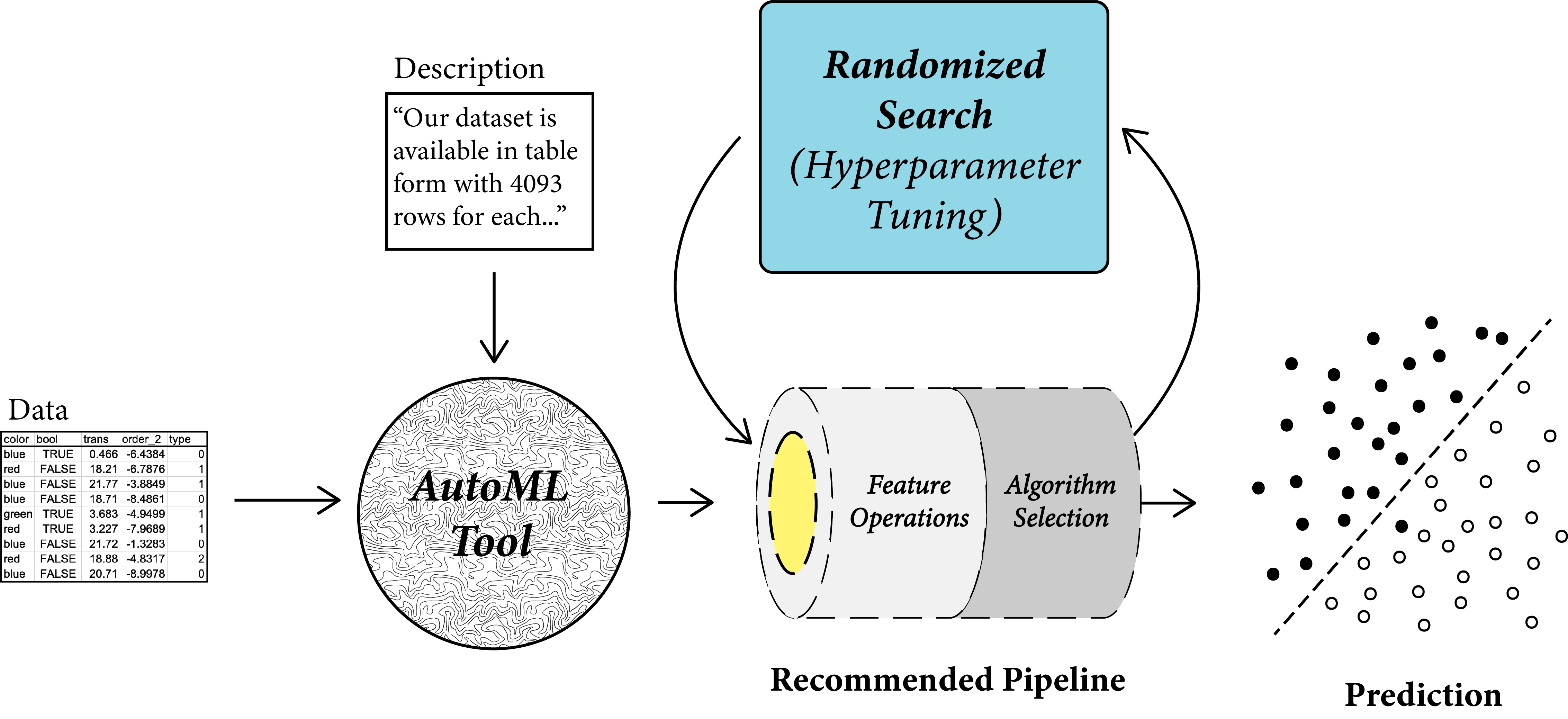}
    \caption{Overview of test procedure, showing how our results are obtained. Our AutoML tool accepts a dataset and a description of this dataset, and outputs a pipeline consisting of an estimator, preceded by a feature processor. We then run a hyperparameter search on the resulting pipeline. Finally, we train and report validation accuracy.}
    \label{fig:automl-general}
\end{figure}

Notice that our method does not need to complete even a single model fit to recommend a model. On the other hand, we must tune hyperparameters and fit the model (to learn the parameters) on the dataset to predict output values for new input data. Our method recommends a model and then tunes hyperparameters by random search, in less than a second. An overview of this process is shown in Fig. \ref{fig:automl-general}.

\section{Results}
\label{sec:results}

\begin{table}[!h]
\small
\centering
\begin{tabular}{l|ccccc|r}
& & & \textbf{Accuracy} & & & \textbf{Time} (seconds)\\
AutoML & Median & Min & Max & Mean & Std. & Median\\
\hline
Autosklearn             &    0.96 &  0.40 &  1.0 &  0.88 &  0.16 &       295.94 \\
Oboe                    &    0.92 &  0.56 &  1.0 &  0.89 &  0.13 &        87.74 \\
OnlyDescription         &    0.77 &  0.10 &  1.0 &  0.69 &  0.27 &         0.16 \\
Zero-Shot (Ours)        &    0.92 &  0.40 &  1.0 &  0.87 &  0.15 &         0.15 \\
Zero-Shot NoDescription &    0.86 &  0.32 &  1.0 &  0.81 &  0.18 &         0.18 \\
\end{tabular}
\caption{Aggregate results, taken across 5 trials. Our zero-shot approach is competitive with Autosklearn and TensorOBOE while being 3 orders of magnitude faster. Autosklearn and TensorOBOE take minutes to provide results, whereas our zero-shot approach takes less than a second.}
\label{tab:results-meanaccuracy}
\end{table}

\begin{figure}[!htb]
    \centering
    \includegraphics[width=\textwidth]{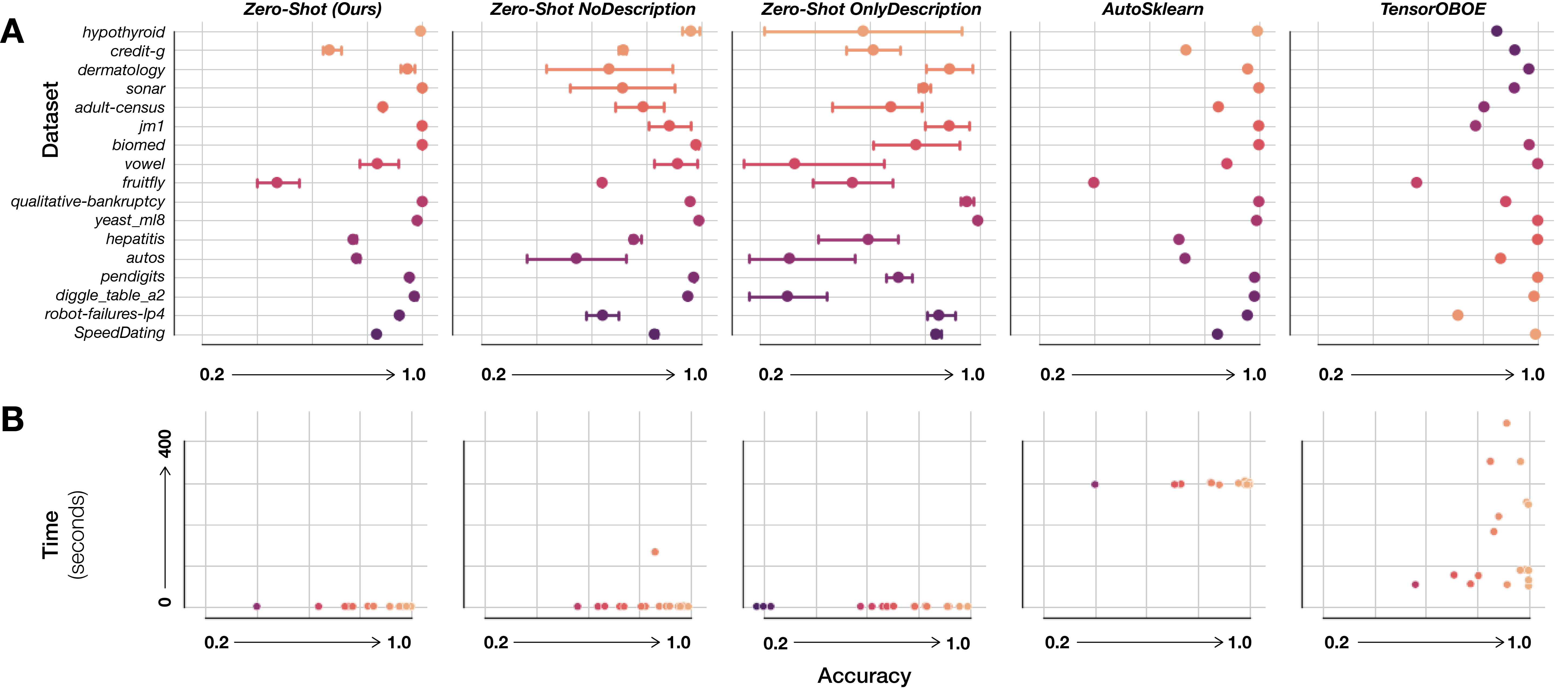}
    \caption{\textbf{(A)} Accuracy per dataset and \textbf{(B)} time vs. accuracy for each tool. Bounds show standard deviations for our zero-shot methods across 5 trials to account for stochastic variation. Overall, we observe consistent time performance even at different levels of accuracy for our tool. AutoSklearn constrains itself to the allotted 5 minutes consistently, whereas TensorOBOE is much less consistent. Our approach achieves similar accuracy levels on several datasets while consistently performing orders of magnitude faster.}
    \label{fig:results-timeaccuracy}
\end{figure}

\begin{table}
\small
\centering
\begin{tabular}{l|ccc|cc}
\textbf{Dataset} &  \textbf{ZS} &  \textbf{ZSND} & \textbf{OnlyDesc.} & \textbf{AutoSk.} & \textbf{Tens.OBOE}\\
\hline
SpeedDating            &              0.83 &                     0.83 &             0.83 &         0.85 &  0.85 \\
adult-census           &              0.85 &                     0.86 &             0.78 &         0.85 &  0.87 \\
autos                  &              0.76 &                     0.56 &             0.17 &         0.73 &  0.80 \\
biomed                 &              1.00 &                     0.98 &             0.62 &         1.00 &  1.00 \\
credit-g               &              0.64 &                     0.72 &             0.66 &         0.74 &  0.71 \\
dermatology            &              0.92 &                     0.88 &             0.92 &         0.96 &  0.99 \\
diggle\_table\_a2        &              0.97 &                     0.95 &             0.23 &         0.98 &  0.97 \\
fruitfly               &              0.40 &                     0.64 &             0.68 &         0.40 &  0.56 \\
hepatitis              &              0.74 &                     0.74 &             0.68 &         0.71 &  0.77 \\
hypothyroid            &              0.99 &                     0.96 &             0.57 &         0.99 &  0.99 \\
jm1                    &              1.00 &                     0.81 &             0.91 &         1.00 &  1.00 \\
pendigits              &              0.95 &                     0.97 &             0.70 &         0.98 &  0.91 \\
qualitative-bankruptcy &              1.00 &                     0.96 &             0.96 &         1.00 &  1.00 \\
robot-failures-lp4     &              0.92 &                     0.67 &             0.83 &         0.96 &  0.92 \\
sonar                  &              1.00 &                     0.90 &             0.79 &         1.00 &  1.00 \\
vowel                  &              0.77 &                     0.92 &             0.20 &         0.88 &  0.88 \\
yeast\_ml8              &              0.98 &                     0.99 &             0.99 &         0.99 &  0.97 \\
\end{tabular}
\caption{Accuracy per test data set (median across 5 trials for the zero-shot methods). \textbf{ZS} is our zero-shot method that combines dataset meta-features with description embeddings, \textbf{ZSND} uses meta-features only (no descriptions), \textbf{OnlyDesc.} only uses the description embeddings (no meta-features). \textbf{AutoSk.} and \textbf{Tens.OBOE} are the existing AutoML systems AutoSklearn and TensorOBOE respectively.}
\label{tab:results-accuracy}
\end{table}

\begin{table}[!h]
\small
\centering
\begin{tabular}{l|ccc|cc}
\textbf{Dataset} &  \textbf{ZS} &  \textbf{ZSND} & \textbf{OnlyDesc.} & \textbf{AutoSk.} &    \textbf{Tens.OBOE}\\
\hline
SpeedDating            &              0.89 &                     0.27 &             1.22 &       300.5 &  351.55 \\
adult-census           &              0.75 &                   132.01 &             0.74 &      299.08 &  181.16 \\
autos                  &              0.45 &                     0.07 &             0.15 &       295.4 &   74.93 \\
biomed                 &              0.66 &                     0.10 &             0.12 &      297.88 &   49.86 \\
credit-g               &              0.11 &                     0.07 &             0.13 &      296.38 &   76.36 \\
dermatology            &              0.14 &                     0.56 &             0.16 &      297.84 &   89.28 \\
diggle\_table\_a2        &              0.14 &                     0.18 &             0.16 &      304.15 &   87.95 \\
fruitfly               &              0.12 &                     0.08 &             0.09 &      295.37 &   53.17 \\
hepatitis              &              0.16 &                     0.07 &             0.10 &      295.43 &   54.58 \\
hypothyroid            &              0.11 &                     0.52 &             0.22 &      295.23 &  253.02 \\
jm1                    &              0.14 &                     0.07 &             0.19 &       300.7 &  246.67 \\
pendigits              &              0.27 &                     1.90 &             0.88 &      294.89 &  443.82 \\
qualitative-bankruptcy &              0.13 &                     0.11 &             0.14 &      294.78 &   87.52 \\
robot-failures-lp4     &              0.11 &                     0.72 &             0.14 &       298.6 &   53.05 \\
sonar                  &              0.13 &                     0.62 &             0.12 &      294.81 &    64.2 \\
vowel                  &              0.15 &                     0.86 &             0.29 &      294.67 &  218.03 \\
yeast\_ml8              &              0.50 &                     0.08 &             0.16 &      295.49 &  350.97 \\
\end{tabular}

\caption{Running time per test dataset in seconds (median across 5 trials for the zero-shot methods). These times are exclusive of embedding and metafeature calculation time, though these are brief: median total (i.e. for embeddings + meta-features extraction) time on the test set is 0.15s (with a mean of 0.196s, standard deviation of 0.15s, minimum of 0.096s, and maximum of 0.73s). Once again, \textbf{ZS} is our zero-shot method that combines dataset meta-features with description embeddings, \textbf{ZSND} uses meta-features only (no descriptions), \textbf{OnlyDesc.} only uses the description embeddings (no meta-features). \textbf{AutoSk.} and \textbf{Tens.Oboe} are the existing AutoML systems AutoSklearn and TensorOBOE respectively.}
\label{tab:results-time}
\end{table}

Table \ref{tab:results-meanaccuracy} shows our results aggregated over a representative set of 17 test datasets, comparing our approach with state-of-the-art AutoML systems and baselines without descriptions (i.e. data meta-features only) and with no data (i.e. description only). Overall, our method results in accuracy comparable to AutoSklearn and TensorOBOE, as well as significant improvements over both baseline models (ND indicates no description, OnlyDescription indicates no data/meta-features) while accelerating the process by orders of magnitude. We additionally report per-dataset accuracy in Table \ref{tab:results-accuracy} and per-dataset time taken in Table \ref{tab:results-time}. Finally, Fig. \ref{fig:results-timeaccuracy} shows the relationship between time taken and accuracy achieved. Our approach achieves similar accuracy levels on several datasets while consistently performing much faster, as is seen in detail in these tables and figures.

We give AutoSklearn and TensorOBOE each a maximum of five minutes to produce a pipeline, with AutoSklearn generally taking about this long and TensorOBOE taking at least 40 seconds and exceeding 5 minutes in the worst case. Our approach is significantly faster without significantly compromising the accuracy of the resulting pipeline. The per-dataset results show that while our method does not always result in the highest accuracy, it often does so by a significant margin over the NoDescription and OnlyDescription versions. Additionally, even the model with only descriptions may result in high accuracy.

\begin{figure}
\begin{lstlisting}[language=Python]
from model.model import AutoMLModel
from model.descriptions import featureops, estimators

model = AutoMLModel.from_pretrained("./{path}.ckpt")
y_featureop, y_estimator = model(X_train) 
featureop = featureops[y_estimator.item()]
estimator = estimators[y_featureop.item()]
pipeline = AutoMLPipeline([featureop, estimator], [X_train, y_train, X_test, y_test])
tuned_pipeline, accuracy, time_tuning, extra_metrics = pipeline()
\end{lstlisting}
\caption{Example code usage: Our AutoML code is modular and easy to use.} 
\end{figure}

\paragraph{Limitations.}
There are several limitations of this work that can easily be addressed in future work. We use 182 datasets (165 train, 17 test) instead of more which our model can easily scale to handle. We used two existing AutoML systems for training and comparison instead of more, though we have conducted initial experiments with more and future work can similarly scale to such systems (including TPOT, AlphaD3M, H2O, etc.). We reported only accuracy instead of F1 scores, though we computed both and found that it does not change our conclusions.

We currently optimize for hyperparameters using random search. Incorporating hyperparameters into our graph representation approach would improve the system. Another limitation is that we only include 34 machine learning primitives (feature extractors, pre-processors, estimators, etc.) though these may be extended. All of these limitations may be alleviated by scaling up the system, in a commercial AutoML system, though our current setup is sufficient for our purposes.

\section{Conclusions}\label{sec:conclusions}
We introduce a new privileged zero-shot approach to AutoML that recommends a good pipeline for a given dataset in real-time. Our system uses textual descriptions of the datasets and machine learning primitives, that typical AutoML systems are not privy to. A graph neural network is then used to predict the best pipeline for a given dataset. Using descriptions, our approach improves performance while being significantly faster than other AutoML systems, reducing running and prediction time from minutes to milliseconds. 

Future work will extend our approach to handle different types of data, including audio and images. In addition, we envision an extension to semi-supervised AutoML by using a GNN to embed a large unsupervised set of datasets without pipelines, together with a small supervised set of datasets with AutoML pipelines. Finally, we make our models and code available in the supplementary material.

\bibliography{main}

\appendix

\end{document}